\documentclass[letterpaper]{article} 
\usepackage{aaai2026}  
\usepackage{times}  
\usepackage{helvet}  
\usepackage{courier}  
\usepackage[hyphens]{url}  
\usepackage{graphicx} 
\usepackage{natbib}  
\usepackage{caption} 
\usepackage{amssymb}
\usepackage{amsmath}
\usepackage{subcaption}
\usepackage{booktabs}
\usepackage{colortbl}
\usepackage{xcolor}
\usepackage{cleveref}
\usepackage{xspace}
\usepackage{bm}

\newcommand{\ie}{\textit{i.e.}\xspace}
\usepackage{algorithm}
\usepackage{algorithmic}

\usepackage{newfloat}
\usepackage{listings}
\DeclareCaptionStyle{ruled}{labelfont=normalfont,labelsep=colon,strut=off} 
\floatstyle{ruled}
\newfloat{listing}{tb}{lst}{}
\floatname{listing}{Listing}
\title{GaussianImage++: Boosted Image Representation and Compression with 2D Gaussian Splatting}
\author{
     Tiantian Li\textsuperscript{\rm 1},
    Xinjie Zhang\textsuperscript{\rm 2,\rm3}, Xingtong Ge\textsuperscript{\rm 3}, Tongda Xu\textsuperscript{\rm 1}, Dailan He\textsuperscript{\rm 4}, Jun Zhang\textsuperscript{\rm 3}, Yan Wang\textsuperscript{\rm 1}\thanks{Corresponding author.}
}
\affiliations{
    \textsuperscript{\rm 1}Institute for AI Industry Research (AIR), Tsinghua University\\
     \textsuperscript{\rm 2}Microsoft Research Asia\\
   \textsuperscript{\rm 3}The Hong Kong University of Science and Technology\\
      \textsuperscript{\rm 4}The Chinese University of Hong Kong\\
      
       litiantian2er@hnu.edu.cn, wangyan@air.tsinghua.edu.cn

}

\usepackage{bibentry}

\begin{document}

\maketitle

\begin{abstract}
Implicit neural representations (INRs) have achieved remarkable success in image representation and compression, but they require substantial training time and memory. Meanwhile, recent 2D Gaussian Splatting (GS) methods (\textit{e.g.}, GaussianImage) offer promising alternatives through efficient primitive-based rendering. However, these methods require excessive Gaussian primitives to maintain high visual fidelity. To exploit the potential of GS-based approaches, we present GaussianImage++, which utilizes limited Gaussian primitives to achieve impressive representation and compression performance. Firstly, we introduce a distortion-driven densification mechanism. It progressively allocates Gaussian primitives according to signal intensity. Secondly, we employ context-aware Gaussian filters for each primitive, which assist in the densification to optimize Gaussian primitives based on varying image content. Thirdly, we integrate attribute-separated learnable scalar quantizers and quantization-aware training, enabling efficient compression of primitive attributes. Experimental results demonstrate the effectiveness of our method. In particular, GaussianImage++ outperforms GaussianImage and INRs-based COIN in representation and compression performance while maintaining real-time decoding and low memory usage.
\end{abstract}

\begin{links}
    \link{Code}{https://github.com/Sweethyh/GaussianImage_plus.git}
\end{links}
\begin{figure}[h]  
    \centering
    \includegraphics[width=\linewidth]{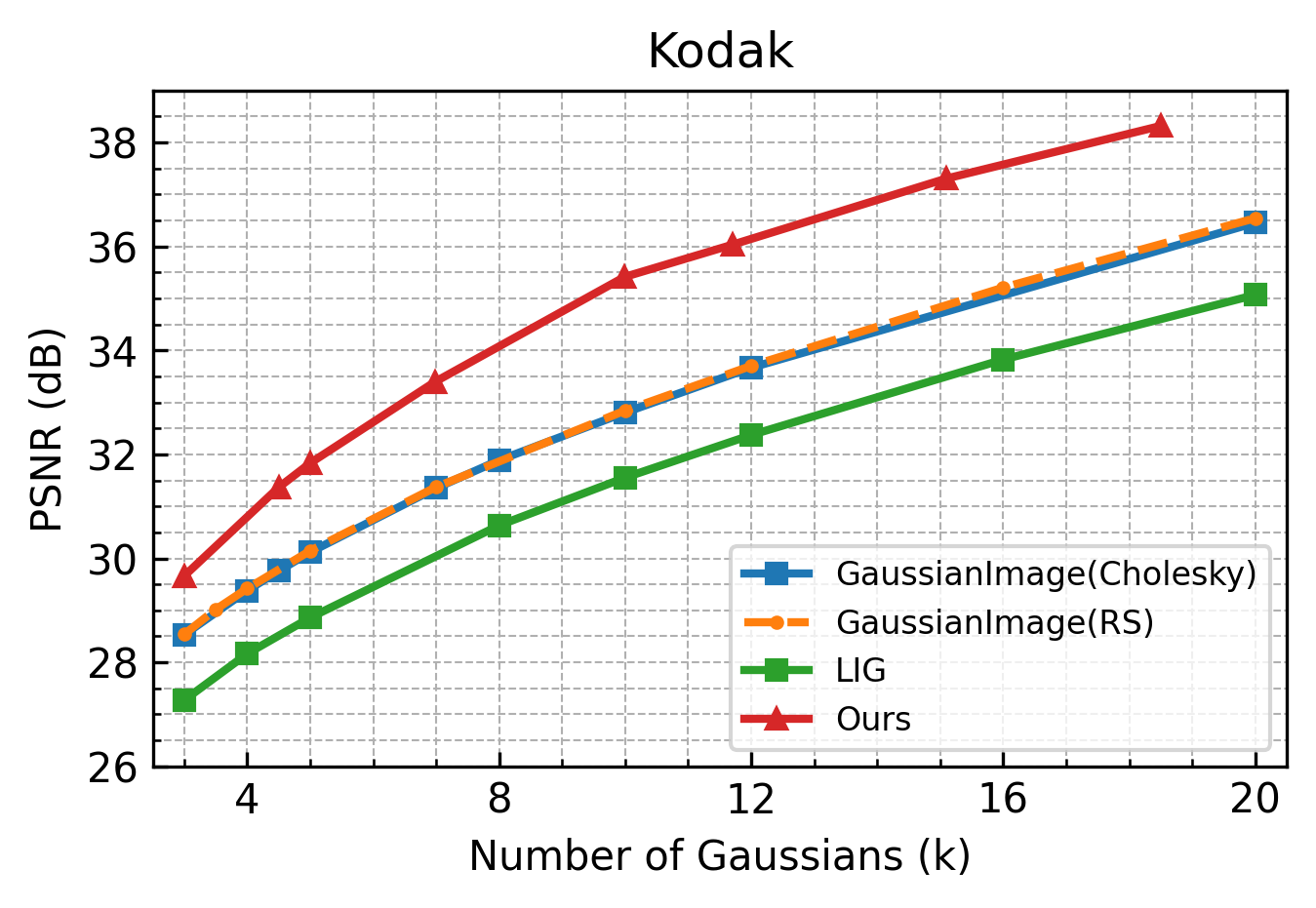}   

    \caption{Image representation based on 2D GS methods with different numbers of Gaussians.
    Our GaussianImage++ exhibits significant performance gains.
    }
    \label{fig:teaser_figure}  

\end{figure}
\section{Introduction}

\label{sec:intro}
Neural image representations and compression have recently emerged as promising techniques for storing, streaming, and rendering visual data. Most neural compression models \cite{balle2018variational,cheng2020learnedloss,he2022elic} are based on autoencoders, with an encoder mapping an image to a high-dimensional but low-resolution latent, a quantizer processing the latent representation for storage and transmission, and a decoder decoding a lossy reconstruction from the quantized latent.

Implicit neural representations (INRs)~\cite{siren,dupont2021coin,dupont2022coin++} have significantly shifted this paradigm. Assisted with lightweight multi-layer perceptrons (MLPs), they have demonstrated remarkable visual fidelity across various applications. However, they suffer from large memory overhead and long training time. Recently,  Gaussian Splatting (GS)~\cite{3dgs} has become popular for tasks ranging from dynamic scene reconstruction~\cite{zhang2025mega,dass} to super-resolution~\cite{peng2025pixel}. GaussianImage~\cite{zhang2024gaussianimage} first applied GS to 2D image representation and compression, significantly reducing memory and training time. However, GaussianImage lacks a densification mechanism to control the number of 2D Gaussians adaptivity based on image content, which affects the full utilization of the representation capabilities, limiting the fitting performance. Mirage~\cite{waczynska2024mirage} represents 2D images using flat 3D Gaussian components, \ie it initializes Gaussians on the XZ plane with the y-axis reduced to zero. It still employs Adaptive Density Control (ADC)~\cite{3dgs} in 3D GS to prune and grow the Gaussians, easily causing uncontrolled growth of the number of Gaussians and leading to out-of-memory errors during training. LIG~\cite{zhu2025large}focuses on fitting large images with numerous 2D Gaussians without compact compression of Gaussian attributes. For storage efficiency, various approaches are tailored for the compression of 3D GS, like LightGaussian~\cite{fanLightGaussianUnbounded3D2023}. In contrast, 2D GS has received less attention. On the one hand, the redundant spherical harmonic coefficient in 3D GS is more suitable for compact Vector Quantization (VQ) compared with color attributes in 2D GS. On the other hand, 3D GS compression methods HAC~\cite{chen2024hac} and ContextGS~\cite{wang2024contextgs} are built upon the neural Gaussians, \ie Scaffold~\cite{lu2024scaffold}, which introduce fundamental differences that prevent their direct application to our 2D GS framework.  Our 2D GS utilizes explicitly stored attributes. In contrast, neural Gaussian approaches assign high-dimensional latent feature vectors to Gaussians, which an MLP then decodes into view-dependent color and opacity.This design is crucial for capturing complex 3D view-dependent appearance and for achieving compression by operating in the latent space. However, it is architecturally mismatched and introduces unnecessary overhead for our goal of efficiently representing a fixed 2D image.

We introduce GaussianImage++ to achieve efficient image representation and compression with 2D GS. Our explorations mainly include progressive training, distortion-driven densification, and context-aware Gaussian low-pass filters. Through these explorations, GaussianImage++ achieves superior performance compared to vanilla 2D GaussianImage on multiple benchmark datasets in image representation and compression while maintaining small disk overhead and fast rendering speeds. Furthermore, our densification and context-aware Gaussian low-pass filters are flexible and universal enhancement techniques for current 2D GS approaches.
Our main contributions are threefold: 
\begin{itemize}
\item We develop a universal boosting technique based on a progressive distortion-driven densification to control the density of 2D Gaussian primitives generated per image.

\item We employ content-aware Gaussian low-pass filters to adaptively adjust the incidence and intensity of each Gaussian primitive, which echoes our densification and improves visual quality.

\item Extensive experimental results demonstrate that our approach achieves significant improvements in various benchmark datasets for image representation and compression. Comprehensive ablation and analysis demonstrate the effectiveness and robustness of proposed components when applied to other 2D GS methods.
\end{itemize}
\section{Related Work}
\label{sec: related work}

\subsection{Explicit Image Representation with 2D Gaussians}
3D Gaussian Splatting represents 3D scenes by a set of explicit Gaussian ellipsoids with learnable attribute parameters.
In 2D scenes, Mirage adapts flat 3D Gaussians within the XZ plane for 2D image editing and representation. GaussianImage pioneers 2D GS for image representation and compression by presenting a compact 2D GS representation and a novel accumulated blending-based rasterization. However, this pioneering work lacks a critical densification process necessary for adaptive Gaussian primitive allocation, and rate-distortion (RD) performance needs to be improved. A follow-up 2D GS work LIG focuses on fitting large images with numerous Gaussians but does not explore the compact compression of  these multi-attribute Gaussians, leading to significant overhead on resource-constrained devices.

\subsection{Neural Image Compression}
Traditional image codecs, like JPEG~\cite{wallace1991jpeg}, JPEG2000~\cite{taubman2002jpeg2000}, BPG~\cite{bpg2014} and so on, feature high visual quality and excellent compression ratio, but complex hand-crafted modules (transformation, quantization and entropy coding) lead to substantial decoding latency and limit real-time applications. Learning-based image compression (LIC) methods~\cite{balle2016end,balle2018variational,he2022elic,cheng2020learnedloss} based on hierarchical variational autoencoders (VAEs) surpass traditional codecs in RD performance through learned nonlinear transforms and entropy models. However, cascaded convolutional blocks and autoregressive entropy coding introduce substantial computational overhead during decoding, limiting their practical deployment.  While efforts toward real-time decoding like EVC~\cite{wang2023evc} improve inference efficiency, they remain grounded in the VAEs/hyperprior framework. GaussianImage++ currently still lags behind these methods in RD performance but offers a solution diagram, at the same time, with significant advantage in decoding speed and memory efficiency.

Image-based INR methods~\cite{siren, dupont2021coin, dupont2022coin++, ladune2023cool,girish2023shacira} provide a novel compression paradigm: An encoder approximates a spatial-to-color mapping via an MLP, quantizes and transmits the MLP weights as compressed codes. A decoder reconstructs the image by querying the MLP at each pixel coordinate.  However, purely relying on implicit MLP architectures leads to prolonged optimization cycles and suboptimal RD performance. Subsequent hybrid approaches C3~\cite{kim2024c3}, COOL-CHIC~\cite{ladune2023cool}, and work~\cite{girish2023shacira}  integrate explicit multi-resolution hash-grid features to accelerate MLP  convergence and enhance visual quality.Nevertheless,  these hybrid models require instance-specific quantizers and autoregressive context entropy models for additional decoding grid latents, introducing substantial computational overhead and decoding time. Diverging from VAEs and INRs paradigms, we employ explicit 2D GS that achieves high visual fidelity, memory efficiency, and real-time decoding speeds.
\begin{figure*}
    \centering
    \includegraphics[width=0.8\linewidth]{
    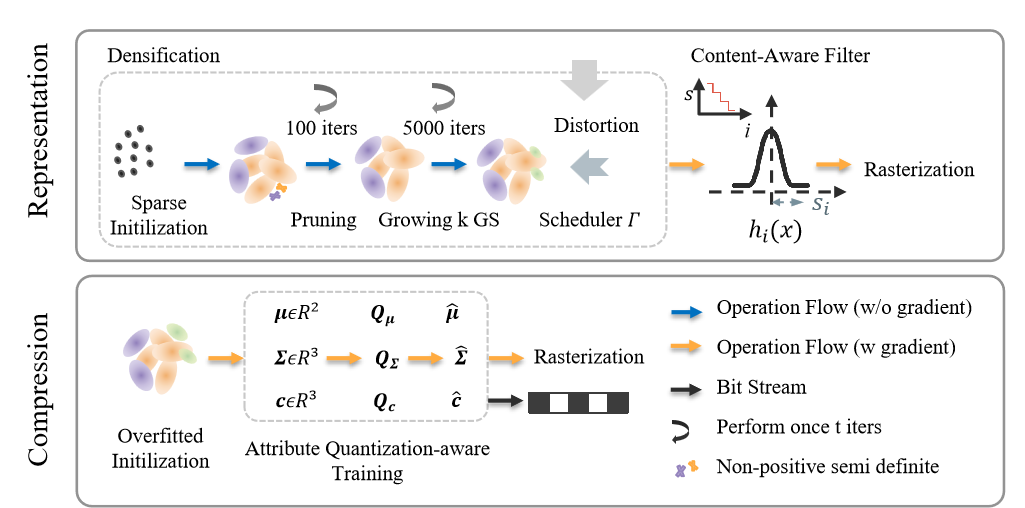}
    \caption{{Our proposed GaussianImage++ framework}. Our representation pipeline uses densification to initialize sparse Gaussians, growing them periodically to improve under-reconstructed areas. A content-aware filter is applied to all Gaussians before accumulated sum rasterization. For compression, we initialize with overfitted 2D Gaussians and employ quantization-aware training to encode attributes into compact bitstreams.
    }
    \label{fig:framework}
\end{figure*}
\section{Methodology}
In this section, we present our  GaussianImage++  framework as illustrated in \Cref{fig:framework}. We begin with the basics of GS and its variants. Subsequently, we detail our two critical enhancements: the distortion-driven densification ($D^3$) and content-aware Gaussian filters (CAF) with adaptive variance. Finally, we discuss our compression techniques.

\label{gs_overview}
\noindent \textbf{2D Gaussian Splatting.}
Motivated by 3D Gaussian Splatting (3DGS), GaussianImage represents 2D scenes using a set of 2D Gaussian ellipticals parameterized by position  $ \bm{\mu}\in \mathbb{R}^2$, covariance $\mathbf {\Sigma} \in \mathbb{R}^{2\times2}$ and color $\mathbf{c}\in \mathbb{R}^3$. Oriented to 2D image representation, GaussianImage eliminates view and project transformation and depth-based Gaussian sorting, and replaces spherical harmonics with RGB colors, $\alpha$ blending with accumulated summation. Consequently, the rendered image  changes to 

\begin{equation}
\label{eq:3DGS}  
G_i(\mathbf{x}) = {\exp(-\frac{(\mathbf{x}- \bm{\mu}_i)^T \mathbf{\Sigma}^{-1} (\mathbf{x}- \bm{\mu}_i)} {2})}      
\end{equation}

\begin{equation}
\label{eq:2DGS}  
\mathbf{C} = \sum_{i \in N} \mathbf{c}_i G_i(\mathbf{x}).
\end{equation}

According to the properties of covariance matrices $\mathbf {\Sigma}$, different parameterization variants exist for 2D GS: 1) Parameterize a rotation matrix $\mathbf {R}$ and scaling matrix $\mathbf {S}$ like 3D GS, where $\mathbf {\Sigma}=\mathbf {R}\mathbf {S}\mathbf {S}^T\mathbf {R}^T$. 2) Parameterize a lower triangular matrix $\mathbf{L}$ like GaussianImage,  leveraging Cholesky factorization  $\mathbf {\Sigma}=\mathbf {L}\mathbf {L}^T$. 3) Directly parameterize  $\mathbf {\Sigma}$ without decomposition like LIG. As stated in LIG, while the positive semi-definite nature of $\mathbf {\Sigma}$ helps Gaussians retain their physical meaning, it is not strictly necessary for image fitting. We also directly parameterize $\mathbf {\Sigma}$. In the experiment, we have proven that our two key components are effective regardless of these three parameterization methods.
\subsection{Distortion Driven Densification }
\label{densification} 
Adaptive Density Control (ADC) in 3D GS removes transparent Gaussians and densifies Gaussians with an average magnitude of view-space position gradients above a threshold. Directly applying ADC in 2D GS is unfeasible since changes accumulated from position gradients are are typically too low to trigger the densification mechanism effectively.Prior 2D GS methods have limitations: GaussianImage lacks progressive densification,  leading to substantial under-reconstructed areas. Although LIG employs a hierarchical approach, which fits a low-scale image and a high-scale residual image, it lacks dynamic Gaussian growing and pruning within each level.

Motivated by these observations, we propose a progressive, distortion-driven densification mechanism, which is a straightforward and image-quality-oriented method with explicit regulation of Gaussians. In summary, our densification mechanism comprises three stages: Sparse initialization, Gaussian growing, and Gaussian pruning. We limit the maximum number of Gaussians to $M$, allowing users to easily trade off rendering quality and memory overhead, especially on devices with limited memory.

\noindent\textbf{Sparse Initialization.} A good initialization is crucial for image quality. 
We employ a random initialization strategy with specific designs. Let  $N_t$ denote the number of Gaussians at iteration iteration $t$. In initialization ($N_0=\frac{M}{2}$), we uniformly and randomly sample Gaussians positions within image coordinates $[0, H)\times[0, W)$, where $H$ and $W$ are image height and width. The covariances $\Sigma$ are randomly and uniformly sampled within the normalized range (0, 1), with diagonal elements constrained by a lower bound of 0.5 to ensure positive semi-definiteness. Colors are initialized as zeros. Sparse initialization also accelerates early training due to the small $N_0$.

\noindent \textbf{Gaussains Growing.} As sparse Gaussians fail to represent high-frequency details, we progressively densify more Gaussians to under-reconstructed regions.  A concurrent densification work \cite{bulo2024revising} in 3D GS re-distributes the per-pixel errors to each Gaussian and uses per-Gaussian error to decide the clone-split of Gaussians.  
 We directly decide the densification based on a per-pixel distortion $D(X,\hat{X})$ between raw image ${X}$ and rendered image $\hat{X}$,  and the current Gaussians allowance  $M-N_t$. We periodically (interval 5000  iterations) perform our densification. In each densification step, we sample $k$ new Gaussian primitives $\Psi:=\{g_1,\ldots,g_k\}$ at pixels with top-$k$ reconstruction distortion.  We parameterize their position attributes as

 \begin{equation}
\label{eq:densify}
   \bm{\mu}_{\Psi} = \xi \left( \operatorname{Top}_k\left(  D({X},{\hat{X}})  \right) \right),
\end{equation}
color attributes  as
\begin{equation}
\label{eq:densify_color}
\mathbf{c}_{\Psi} = X\left( \xi \left( \operatorname{Top}_k\left(  D(X, \hat{X})  \right) \right) \right).
\end{equation}
The covariance attributes $\mathbf{\Sigma}_{\Psi}$ are parameterized with the same method as the initialization stage. Among $k$ is a variable decided by a scheduler $\tau(t, N_t, M) = \frac{M-N_t}{2}$, $\xi$ is an index function that returns the two-dimensional coordinates of pixels with top-k distortion. $D()$ is an arbitrary image quality distortion function. We use L1 loss for computational simplicity. 

\noindent\textbf{Gaussians Pruning.} Directly optimizing covariance matirx $\mathbf{\Sigma}$ may violate the positive semidefinite in mathematics, which makes \Cref{eq:3DGS} not work. Consequently,  these invalid Gaussians make no contributions to rendering. We prune them to save memory overhead and allowance of Gaussians. As shown in \Cref{fig:framework}, we periodically (100 iterations) check the semi-positive quality of $\mathbf{\Sigma}$ and prune invalid ones for saving memory overhead and allowance of Gaussians. 

\begin{figure}[h]
    \centering\includegraphics[width=\linewidth]{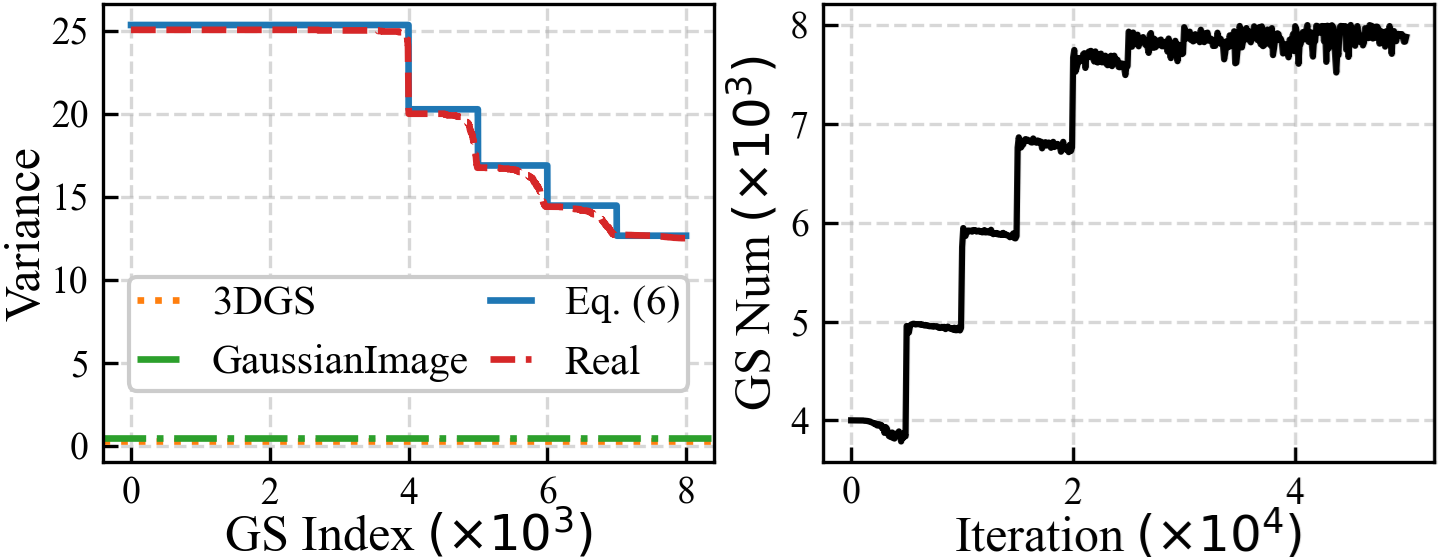}
    \caption{Left: the variance $s_i$ of the i-th Gaussians primitives. Right: Number of Gaussians $N_t$ at iteration $t$.
}
    \label{fig:scale_curve}
\end{figure}
\subsection{Content Aware Gaussian Filters} 
\noindent3D GS  commonly uses elliptical weighted average (EWA) splatting \cite{zwicker2002ewa}  for anti-aliasing before tile-based rasterization. This addresses sampling rate variations caused by perspective projection from 3D Gaussian kernels to 2D screen-space splats. Such anti-aliasing is achieved by applying a resampling filter to the projected 2D Gaussian splats. Consequently, the original footprint function $G_i(x)$ is modified to
\begin{equation}
G_i'(\mathbf{x}) = e^{-\frac{1}{2}(\mathbf{x} -  \bm{\mu}_i)^T (\mathbf{\Sigma}_i+ \mathbf{s}I)^{-1} (\mathbf{x} -  \bm{\mu}_i)}.
\end{equation}
In essence, this applies a zero-mean Gaussian low-pass filterr $h(x)$  to the original Gaussian reconstruction kernel $G_i(\mathbf{x})$, \ie, $G_i'(\mathbf{x})=(G_i\otimes h)(x)$, where $s$ is a scalar representing the variance of  $h(x)$ and $I$ is a unit matrix. 
3D GS and GaussianImage use a constant $s$, 0.3 and 0.5, respectively. In GaussianImage++, we propose a content-aware filter $h(x)$ governed by an adaptive variance vector $\mathbf{s} \in \mathbf{R}^{N_t}$. Each element $s_i \in \mathbb{R}_+$ ($i = 1,\dots,N_t$) 
controls the filter strength for the \textit{i}-th Gaussians and is derived by
\begin{equation}
\label{eq:scale_of_filter}
s_i=\left\{\begin{matrix}
  &\frac{HW}{\alpha}N_t, & i>N_{t-1} \\    
  &s_{i-1}, & i \leq N_{t-1} \\
\end{matrix}\right.
\end{equation}
where $\alpha$ is a scaling factor. \Cref{fig:scale_curve} plots theoretical values derived from \Cref{eq:scale_of_filter} and real values of $\mathbf{s}$ observed during training. Real $\mathbf{s}$ fluctuates slightly due to pruning. Importantly, $\mathbf{s}$ does not consume additional storage, as we store the filtered covariance $\mathbf{\Sigma} + \mathbf{s}I$.

Our CAF adaptively adjusts the sampling rate of the pixel grid by scaling the intersection area of Gaussians and pixels, effectively reducing holes and artifacts in the rendered image. As shown in \Cref{fig:scale_curve} (left), for the early GS, 
we impose a strong filter $h(x)$ with large $s_i$. The large $s_i$ enlarges the original 2D Gaussian reconstruction kernel, which significantly magnifies their coverage area and effectively mitigates undersampling and reduces large ``holes'' in the rendered image. This strategy enables early-stage rendering to produce a recognizable, coarse image, crucial for guiding optimization. As densification begins, we gradually weaken the filter strength, \ie reducing $s_i$ variance for newly added Gaussian primitives. This prevents the filter from dominating the reconstruction kernel, allowing new GS to focus on finer details.

\begin{figure*}[h]
    \centering
\includegraphics[width=0.9\linewidth]{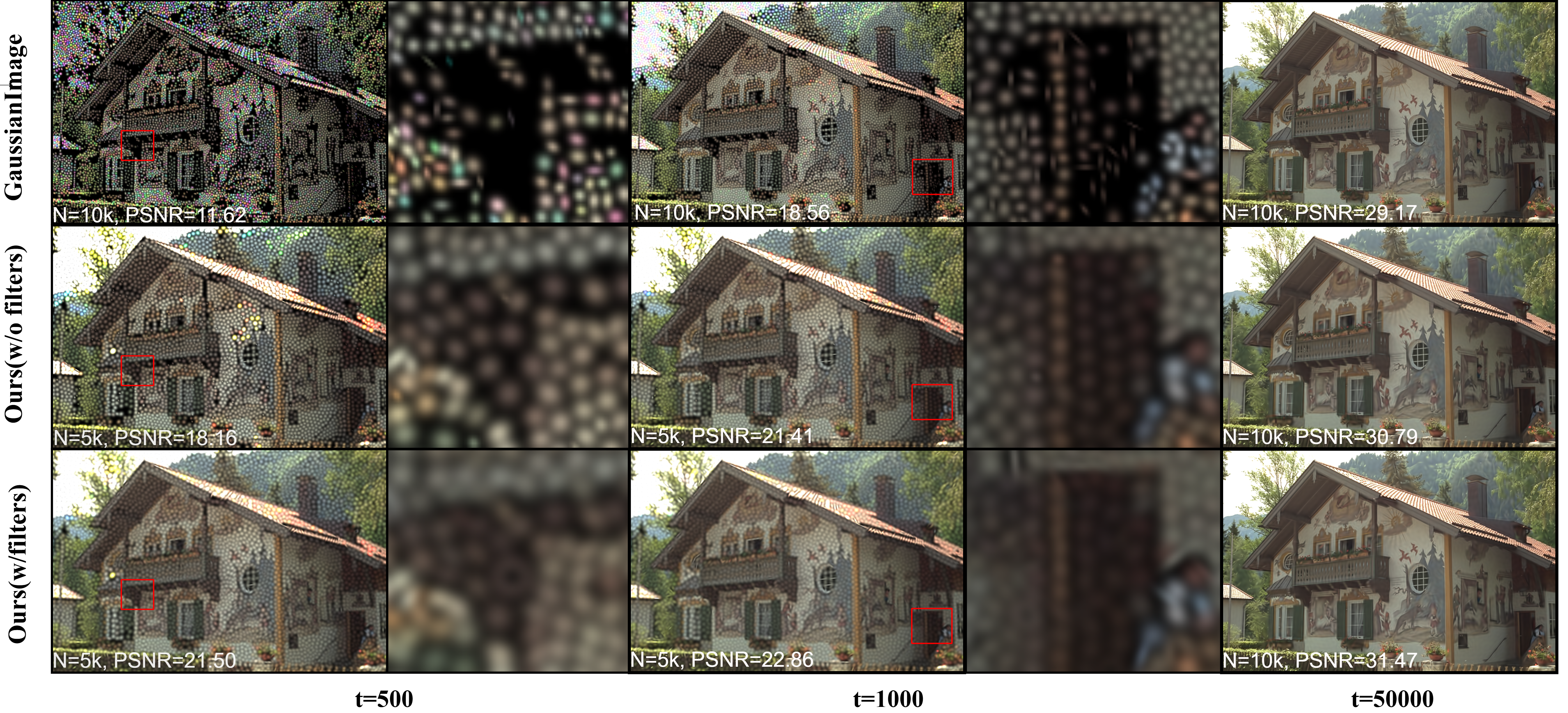}
    \caption{{Progressive training of GaussianImage++ ($M=10k$, $T=5\times 10^4$).} Before densification (t=500), GaussianImage++ with CAF enables reducing the holes and artifacts and represent a coarse structure with sparse Gaussians. As densification steps (t=1000 or t=50000), the final rendered image shows enhanced visual quality.}
    \label{fig:filters}
\end{figure*}

Moreover, the adaptive filtering mechanism aligns with our progressive densification strategy to assist in Gaussian optimization for distinct image content. As shown in \Cref{fig:filters}, at early training stages ($t=500$ and $t=1000$), both GaussianImage and our method without filters (first and second rows) inevitably exhibit significant artifacts and large holes due to $N_t\ll HW$ (the regions are marked with red squares and zoomed into the next patches), since some pixels without any GS intersected. In contrast, the third row in \Cref{fig:filters}, GaussianImage++ with  CAF,  renders a smooth and recognizable image plane even at $t=500$ and $N_t \approx \frac{M}{2}$. This early coarse rendering capability ultimately contributes to better final visual quality, as later Gaussians can focus on refining details more effectively.

\begin{figure*}[h]
    \centering
\includegraphics[width=0.8\linewidth]{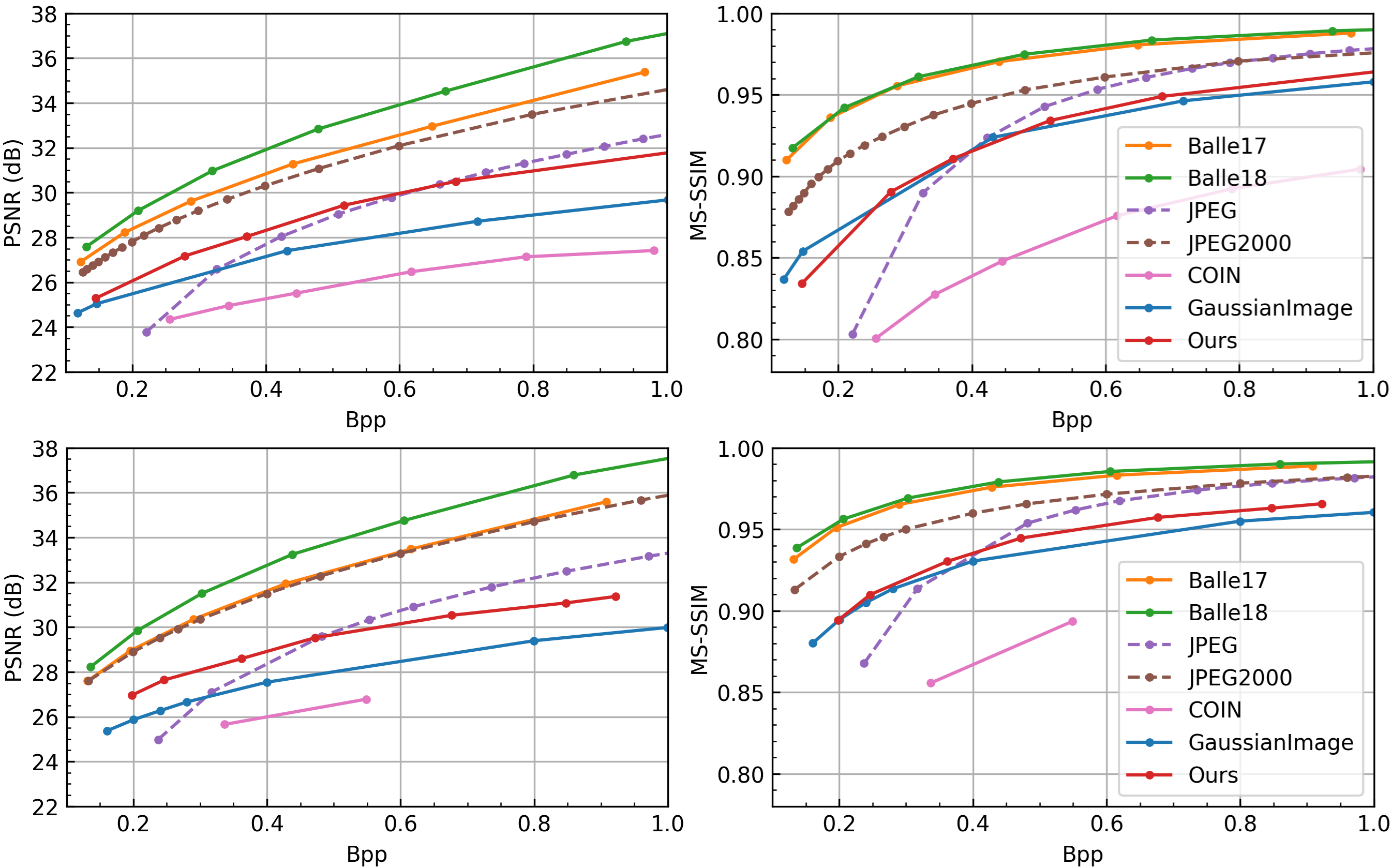}
    \caption{Rate-distortion curves of GaussianImage++ and different baselines on Kodak (top) and DIV2K (bottom) datasets.}
      \label{fig:rd_performance}
\end{figure*}

        
\subsection{Compression Framework}
As shown in \Cref{fig:framework}, we first perform the image representation pipeline with certain steps,  enabling the Gaussians with overfitted initial attributes for quantization. Secondly,
we conduct attribute quantization-aware training to fine-tune Gaussians with more compact attributes. Given a collection of 2D Gaussian primitives, we apply distinct bit-depth LSQ+ \cite{bhalgat2020lsq+} quantizers to different attributes.  LSQ+ is an effective low-bit quantization technique through learnable offsets  $\beta$ and scale $s$. For a vector $\textbf{v}$,  it is quantized and de-quantized with gradient flow by 
\begin{equation}
    \mathbf{\overline{v}}=\lfloor \operatorname{clip}(\frac{\mathbf{v}-\beta}{s},0,2^{b}-1) \rfloor, \quad \mathbf{\hat{v}}=\mathbf{\overline{v}}*s+\beta.
\end{equation}
Specifically, since the sensitivity of geometry attributes ($ \bm{\mu}$ and $\mathbf{\Sigma}$), we adopt 12 and 10-bit quantization precision for $ \bm{\mu}$ and $\mathbf{\Sigma}$. For color attributes, we adopt 6-bit quantization precision.
Such a bit-depth configuration is to balance bitrate and image quality. More comparative analysis and ablation studies are detailed in the supplementary materials.

\section{Experiments}
\subsection{Experimental Setup}
\textbf{Datasets.} We evaluate image representation and compression on two popular datasets: Kodak~\cite{kodak1993kodak} (24 images, $768\times512$ resolution) and the DIV2K HR validation set~\cite{div2k} (100 images, 2K resolution).

\noindent \textbf{Evaluation Metrics}. We use PSNR and MS-SSIM to quantify visual fidelity, comparing GaussianImage++ with GS-based state-of-the-art and INR-based
counterparts. We also report parameter size, training time and rendering speed in the image representation task. For image compression, we use rate-distortion performance and encoding/decoding time.

\noindent \textbf{Implementation Details.} GaussianImage++  is developed on top of GaussianImage with 8 trainable parameters as noted in \Cref{eq:2DGS}, among covariance $\Sigma$ without decomposition unless otherwise stated. We optimize these parameters for 50,000 iterations using the Adam optimizer and L2 loss between ground truth and rendered images. Learning rates for all attributes are initialized at 0.18, halving to 0.5 after 20,000 iterations.  In compression experiments, we perform warm-up training (6000 iterations) before the attribute quantization-aware fine-tuning. Quantizer learning rates are 0.001, decaying by half after 20,000 iterations. Experiments are conducted using NVIDIA A30 GPUs and PyTorch.

\subsection{ Image Representation}
\textbf{Comparison Results}. 
 We compare our method with GS-based methods (3D GS, GaussianImage, LIG and Mirage) and INR-based methods like Siren~\cite{siren}. \Cref{table:reprentation_effiency} presents quantitative results including PSNR, MS-SSIM, Parameters Number, GPU Memory, Training and Rendering time. For fair comparison, all GS-based methods are set to the same maximum numbers of GS. It is clear that compared with GS-based methods, INRs like Siren suffer from large training memory and parameter size, long encoding time and lower FPS. Compared with 2D GS methods, GaussianImage++ significantly enhances visual quality only at the expense of slightly longer training time, since periodic GS densification operation. Fortunately, our rendering speed is not affected by these operations and even exceeds the GaussianImage.  Furthermore, as shown in \Cref{fig:teaser_figure}, our method consistently outperforms previous 2D GS approaches across all the gaussian number range.

\subsection{Image Compression}
\textbf{RD Performance.} \Cref{fig:rd_performance} presents the RD curves across
Kodak and DIV2K datasets.Our codec outperforms INR-based COIN\cite{dupont2021coin}  regardless of PSNR and MS-SSIM  across all datasets. Compared to the GaussianImage baseline, GaussianImage++ surpasses it at the majority of bit-rate points across datasets and metrics, bringing significant improvement regarding bpp-PSNR.
Furthermore,  our method demonstratesnotable advantages over conventional JPEG codec at low bitrates (0.1-0.7bpp in Kodak and 0.1-0.5bpp in DIV2K). In high bitrate points, GaussianImage++ shows comparatively inferior performance. For learnable image codecs, all codecs based on 2D GS still have obvious performance gaps.

\noindent\textbf{Computational Efficiency.} \Cref{table:codec_effiency} reports the computational efficiency of various codecs on Kodak and DIV2K datasets at low/high bitrate. Unlike the image representation, the integration of attribute quantization-aware training  prolongs the encoding times of GaussianImage and GaussianImage++. The conventional and learnable codec still has faster encoding speeds, which remains a singnifcant rooms for GS-based methods. GS-based methods display the most impressive decoding speed, especially for GaussianImage++, maintaining real-time decoding superiority over conventional codec JPEG and learning-based codecs. See supplementary materials for more quality/efficiency comparisons with other state-of-the-art codecs.

\begin{table*}[htb]
\centering
\small
\resizebox{\textwidth}{!}{

\begin{tabular}{@{}lcccccccccccc@{}}
\toprule
 & \multicolumn{6}{c}{Kodak} & \multicolumn{6}{c}{DIV2K} \\
 \cmidrule(lr){2-7} \cmidrule(lr){8-13}
Methods & PSNR & MS-SSIM & Params(M) & GPU MEM(MiB) & Training/s & Render/fps &
PSNR & MS-SSIM & Params& GPU MEM & Training & Render \\
\midrule
\textbf{INR-based} \\
Siren & 26.50 & 0.875 & 3.74 & 2044 & 889.45 & 977 & 29.00 & 0.886 & 0.42 & 23913 & 4957 & --\\
\midrule
\textbf{GS-based} 
& \multicolumn{6}{c}{Number of GS (5k)} & \multicolumn{6}{c}{Number of GS (10k)} \\
\cmidrule(lr){2-7} \cmidrule(lr){8-13}
3D GS & 27.02 & 0.942 & 0.30 & 922 & 331.39 & 988 & 24.60 & 0.897 & 0.59 & 2652 & 1287.44 & 458\\
MiraGe$\dagger$ & 29.27 & 0.958 & 0.10 & 1242 & 247.93 & 1695 &
\cellcolor{yellow!25}27.32 & \cellcolor{red!25}0.921 & 0.25 & 2802 & 2146 & 584\\
GaussianImage & \cellcolor{yellow!25}29.85 & \cellcolor{red!25}0.962 & 0.04 & 812 & 93.49 & 2305 & 26.54 & 0.910 & 0.08 & 1042 & 275.49 & 825\\
LIG & 28.25 & 0.946 & 0.04 & 832 & 118.40 & 1394 & 24.66 & 0.848 & 0.08 & 964 & 288.57 & 834\\
Ours & \cellcolor{red!25}31.83 & \cellcolor{yellow!25}0.960 & 0.04 & 856 & 116.05 & 2365 &
\cellcolor{red!25}28.14 & \cellcolor{yellow!25}0.914 & 0.08 & 958 & 323.74 & 1011\\
\midrule
& \multicolumn{6}{c}{Number of GS (10k)} & \multicolumn{6}{c}{Number of GS (50k)} \\
\cmidrule(lr){2-7} \cmidrule(lr){8-13}
3D GS & 29.99 & 0.968 & 0.59 & 918 & 337.67 & 947 & 29.58 & 0.965 & 2.95 & 3298 & 1422.47 & 351\\
MiraGe$\dagger$ & 30.41 & 0.961 & 0.21 & 1234 & 387.71 & 496 &
\cellcolor{yellow!25}33.23 & \cellcolor{red!25}0.981 & 1.16 & 2268 & 2153.88 & 419\\
GaussianImage & \cellcolor{yellow!25}32.48 & \cellcolor{yellow!25}0.982 & 0.08 & 814 & 115.89 & 2009 & 31.45 & 0.977 & 0.40 & 1062 & 192.61 & 662\\
LIG & 31.00 & 0.975 & 0.08 & 832 & 125.37 & 1331 & 29.71 & 0.965 & 0.40 & 1085 & 294.32 & 620\\
Ours & \cellcolor{red!25}35.41 & \cellcolor{red!25}0.983 & 0.08 & 876 & 118.76 & 2216 &
\cellcolor{red!25}33.75 & \cellcolor{yellow!25}0.978 & 0.38 & 962 & 356.71 & 765\\
\bottomrule
\end{tabular}
    }
\caption{Quantitative evaluation across various baselines on Kodak and DIV2K datasets. 5k represents the number of GS primitives. Mirage$\dagger$ represents that we early stop ADC in Mirage when the current numbers of Gaussians surpass the set maximum number $M$.}

\label{table:reprentation_effiency}
\end{table*}
\begin{figure}[htb]
    \centering
    \includegraphics[width=\linewidth]{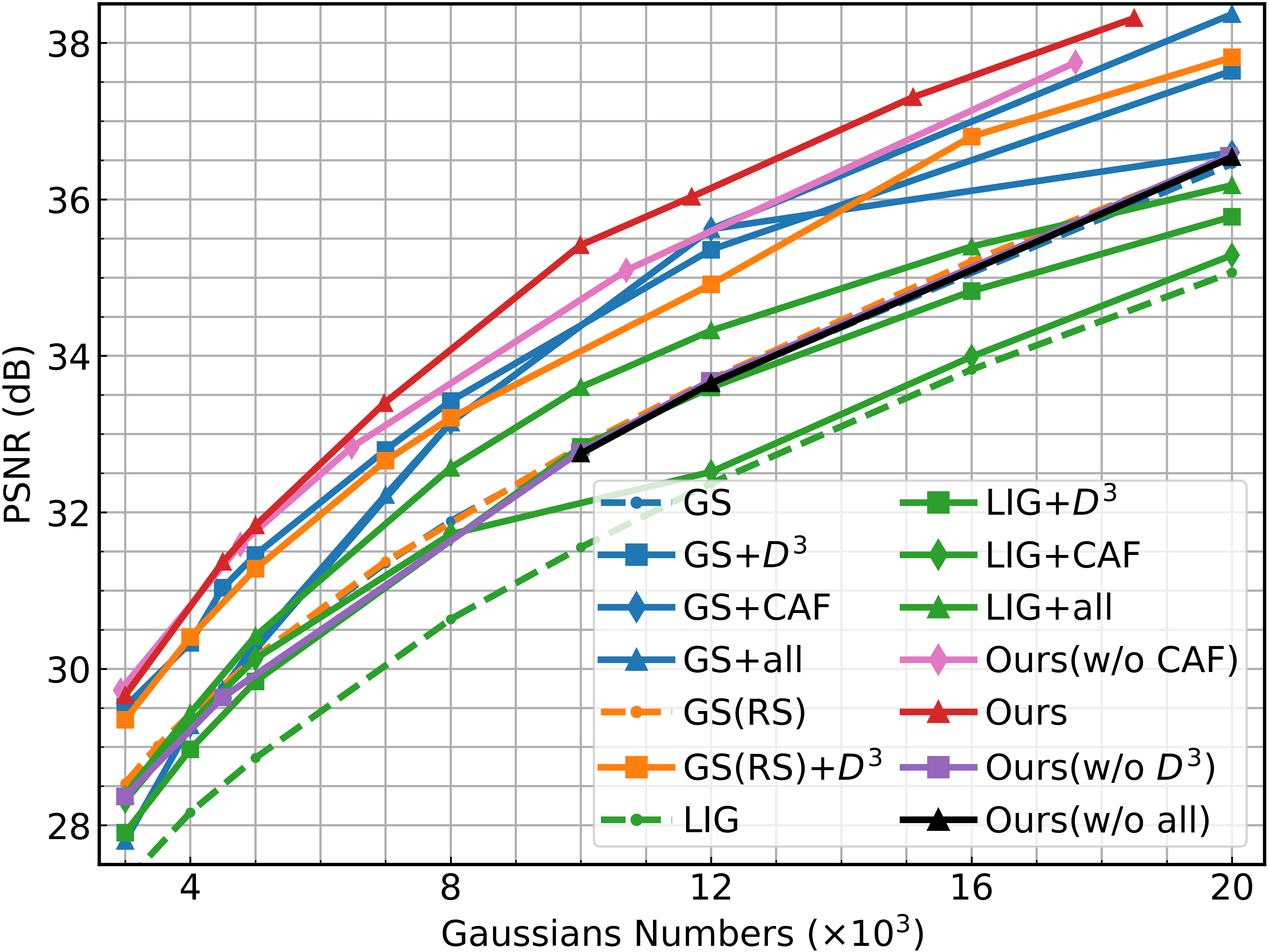}  
    \caption{Ablation of components : distortion-driven densification ($D^3$) and content-aware Gaussian filters (CAF) across 2D GS-based image representation methods on Kodak.}
    \label{fig:ablation}
\end{figure}

\begin{table*}[htb]

\label{table:codec_effiency}
 \small
\resizebox{\textwidth}{!}{
\begin{tabular}{@{}lcccccccccc@{}}
\toprule
 & \multicolumn{5}{c}{Kodak} & \multicolumn{5}{c}{DIV2K} \\ \cmidrule(lr){2-6} \cmidrule(lr){7-11}
 & Bpp & PSNR & MS-SSIM & Encode/s & Decode/FPS & Bpp & PSNR & MS-SSIM & Encode/s & Decode/FPS \\ \midrule
JPEG & 0.22/1.03 & 23.8/32.8 & 0.803/0.979 & 0.012/0.014 & 377/148 & 0.24/0.85 & 25.0/32.5 & 0.867/0.978 & 0.011/0.014 & 188/144 \\
JPEG2K & 0.13/1.19 & 26.6/35.7 &0.882/0.981&    0.582/0.419&    2/3 & 0.27/0.96 & 29.9/25.7 &0.945/0.982 &  0.533/0.554&4/5   \\
Ball\'e17 & 0.12/0.97 & 26.9/35.4 & 0.910/0.988 & 0.029/0.049 & 63/32 & 0.29/0.91 & 30.3/35.6 & 0.965/0.989 & 0.216/0.908 & 6/6  \\
Ball\'e18 & 0.13/0.94 & 27.5/36.7 & 0.917/0.989 & 0.034/0.051 & 44/28  & 0.30/0.86 & 31.6/38.8 & 0.913/0.965 & 0.237/0.385 & 7/7  \\
COIN & 0.17/0.98 & 24.9/27.4 & 0.827/0.904 & 457/726 & 769/344 & 0.27/1.33 & 26.6/29.0 & 0.856/0.886 & 4487/4957 & 11/22  \\
GaussianImage & 0.15/1.00 & 25.0/29.7& 0.854/0.958 & 293/376 & 1827/1822& 0.28/1.00 & 26.7/29.3 & 0.914/0.960 & 376/396 & 857/723  \\
Ours & 0.15/1.08 & 25.3/31.1 & 0.834/0.961 & 338/347 & 1839/1666 & 0.25/0.92&   27.6/31.4&  0.910/0.966 &488/576    &440/748  \\ \bottomrule
\end{tabular}
}
\caption{Computational efficiency evaluation of various image codecs
on two Datasets at low and high Bpp. Traditional codecs are run on a 12th Gen Intel(R) Core(TM) i5-12600K 3.70 GHz processor. Other codecs are operated on an NVIDIA A30 GPU.}
\label{table:codec_effiency}
\end{table*}

\begin{table}[htb]
\centering
\resizebox{\linewidth}{!}{
\begin{tabular}{l|cc|cc}
\toprule
Variants  &BD-PSNR(dB)$\uparrow \quad$& BD-Rate(\%)$\downarrow$ & BD-MS-SSIM$\uparrow \quad$ &BD-Rate \\ \midrule

LSQ+/LSQ+     & 0    & 0          & 0          & 0        \\ 
FP16/LSQ+ &{-0.761}     & 25.11      & {-0.009}     & 17.77    \\ 
FP16/RVQ & {-2.471}     & 138.88     & {-0.036}     & 89.75    \\ 
LSQ+/RVQ  & {-2.491}     & 147.24    &{-0.030}     & 68.29    \\ \bottomrule
\end{tabular}
}
\caption{Ablation of quantization strategies. The first is our final strategy. The left and right of ``/'' respectively indicate the quantization strategies of position/color attributes.}
\label{table:ablation_quant}
\end{table}
\subsection{Ablation Study}
\noindent \textbf{Effect of Different Components.} To highlight the contributions of the key
components distortion-driven densification ($D^3$) and content aware Gaussian filters (CAF), we conduct comprehensive ablation studies. These components are selectively integrated into existing 2D GS methods and our approach, all sharing identical training settings for fair comparison. \textbf{GS (Cholesky)} and \textbf{GS (RS)}  refer to GaussianImage methods using Cholesky and RS factorization for 2D covariance decomposition, respectively. \textbf{LIG} and \textbf{Our(w/o $\mathbf{D^3}$ CAF)} directly optimize learnable covariance without decomposition. As illustrated in \Cref{fig:ablation}, our two components consistently enhance representation performance regardless of the specific optimization strategy. In particular, the $D^3$ component yields the most significant quality enhancement - up to approximate 2dB PSNR gains over the GS (Cholesky) with the same number of Gaussians. And alone $D^3$ also enhances baseline performance, especially for the case of fewer Gaussians. Combining $D^3$ with  CAF, our GaussianImage++ attains further enhancements - up to 3dB PSNR gains over GaussianImage and 4dB PSNR gains over LIG.

\noindent \textbf{Effect of Different Attributes Quantizers.}
\Cref{table:ablation_quant} investigates the effect of different attribute quantization strategies on image compression. We alternatively replace original quantizers (LSQ+) of position/color attributes with Residual Vector Quantization (RVQ) and FP16 (the quantization in GaussianImagae). In this context, ``FP16/LSQ+'' denotes that positions use FP16 quantization while colors use LSQ+, and so forth. The results show significant performance degradation when substituting our LSQ+ position/color quantization modules with other approaches.  For the first two rows, learnable LSQ+ quantization of position proves more bit-efficient and slightly improves image quality, since quantization-aware training can supervise the Gaussians to adjust their attributes to adapt to the quantization errors. From the last two rows, RVQ for color leads to obvious performance drops due to the limited representation capability of its codebook.
Moreover, we ablate the bit-depth configuration for our quantization strategies in supplementary material.

\section{Conclusion}
In this work, we introduce GaussianImage++, a boosted 2D Gaussian Splatting paradigm for image representation and compression. It achieves significant improvements in visual quality and rate-distortion performance while maintaining real-time rendering speeds. These advancements are driven by two novel components: 1) A distortion-driven densification progressively densifies  Gaussians to the image plane, which structurally reduces the number of Gaussian primitives and achieves a better trade-off between the Gaussian compactivity and rendering quality. 2) The content-aware Gaussian filters, in conjunction with our densification, assist in the 2D GS progressive fitting of distinct details. These two components address limitations in existing 2D GS methods. Furthermore, they are generalizable techniques, which can be friendly integrated with other 2D GS methods for boosted performance.

\section{Limitations}
GaussianImage++ still currently lags behind state-of-the-art neural image codecs, especially in high bitrates. Bridging this gap requires exploring more advanced attribute encoding schemes and entropy models. Furthermore, current encoding time is far from realtime, requiring further optimizing the training and quantization processes.

\section{Acknowledgments}
This work was supported by Wuxi Research Institute of Applied Technologies, Tsinghua University under Grant 20242001120.
\bibliography{aaai2026}

\newpage
\appendix
\clearpage
\appendix
\setcounter{page}{1}

\begin{figure*}[!htb]     
    \centering
\includegraphics[width=0.9\textwidth]{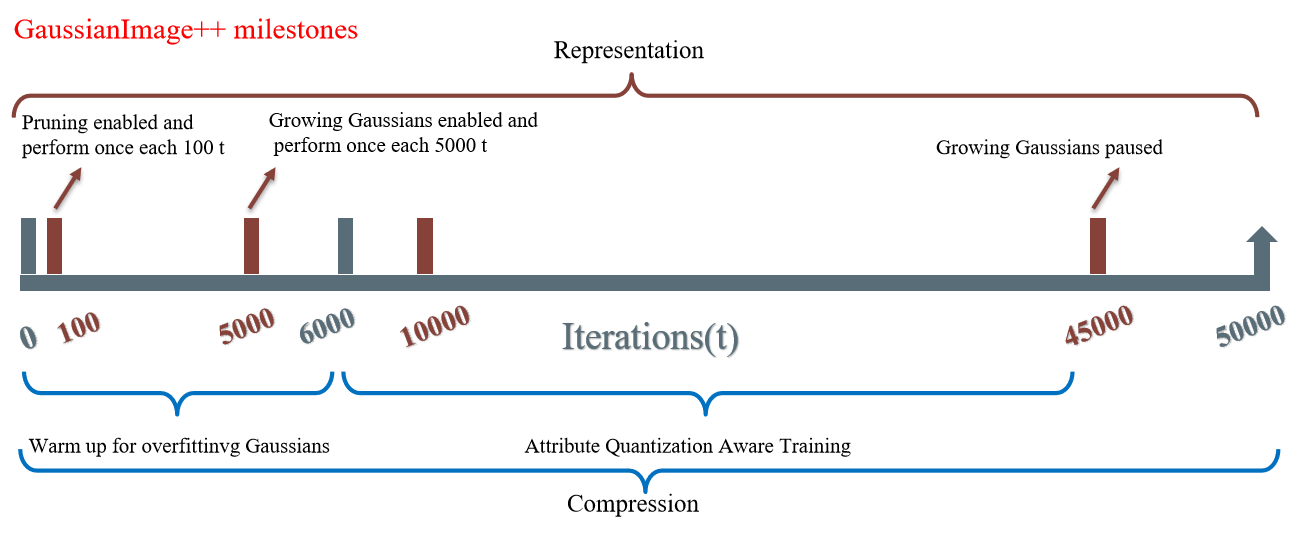}
    \caption{Detailed training process of our 
GaussianImage++. We use the \textcolor[HTML]{864139}{red line} to indicate the training process of the representation task while using the \textcolor[HTML]{0070C0}{blue line} for compression.}
\label{fig:milestone}
\end{figure*}
\section{Appendix}

\begin{figure*}[htb]
    \centering
    \includegraphics[width=0.9\textwidth]{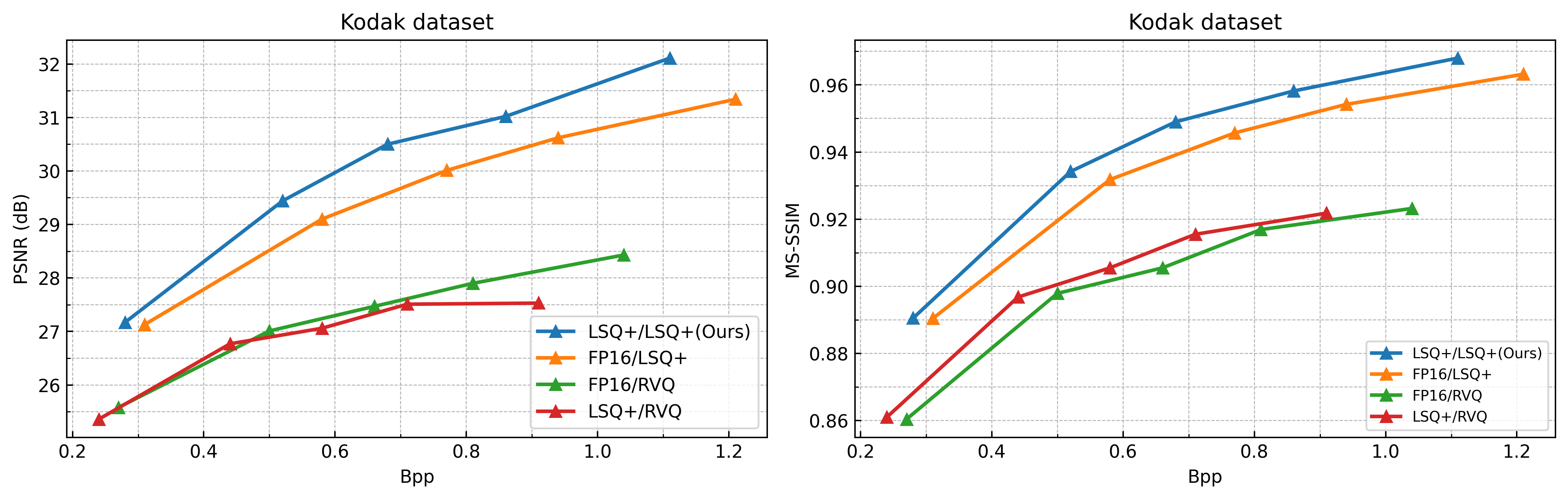}
    \caption{Image compression results with different quantization strategies for position/color attributes on the Kodak dataset.
    }
\label{fig:supple_ablation}  
\end{figure*}
\subsection{Implementation Details}

\noindent\textbf{Training GaussianImage++}.We provide a detailed overview of the training process (total iterations $T$=50,000) for our GaussianImage++ framework, as illustrated in \Cref{fig:milestone}. For the representation pipeline, we perform the positive semi-definite pruning by checking the covariance whether it meets the necessary and sufficient conditions: $det(\mathbf{\Sigma}) \geq 0 $ and $\mathbf{\Sigma}_{11}  \geq 0,\mathbf{\Sigma}_{22}  \geq 0 $. After $t>=5,000$, we grow $k$ Gaussians every 5000 iterations until the $t=45,000$. For the compression pipeline, we first perform the image representation pipeline for warming up until $t=6,000$ to enable the Gaussians with overfitted initial attributes for quantization training. Secondly, we conduct attribute quantization-aware training to fine-tune Gaussians with more compact attributes. During the attribute quantization-aware training, we still retain the pruning operation but pause the Gaussians growing.

\noindent\textbf{Benchmark.} We provide a detailed explanation about the evaluation of other benchmark methods as follows. 
\begin{itemize}
    \item \textbf{INR-based Method}. For Siren~\cite{siren} and Coin~\cite{dupont2021coin}, we use their official codes. For all datasets, we use the full resolution for evaluation. 

\item \textbf{3D-GS methods.}
The performance of 3D GS is evaluated upon gsplat, an open-source library for CUDA-accelerated rasterization of Gaussian. The evaluation of Mirage is based on the official codes offered. In particular, we modify the Mirage by early stopping the ADC densification if the current number of Gaussians surpasses the maximum limited number $M$. So the final number of Gaussians will be larger than $M$.We fit the INR model on Kadak datasets for 50,000 iterations and 20,000 iterations for the DIV2K dataset.
\item \textbf{Traditional 
Codecs and Learnable Image Codecs}
JPEG~\cite{wallace1991jpeg} is evaluated by JPEG. Pillow version 7.1.2.
JEPG2000~\cite{taubman2002jpeg2000} is evaluated by the reference software OpenJpeg v2.5.0. 
We evaluate the VAE-based codecs Ballé17 and Ballé18 using the MSE-optimized models with CUDA provided by CompressAI~\cite{begaint2020compressai}.
\end{itemize}

\noindent\textbf{Components ablation study setting.} In the main text, we have demonstrated the robustness of our components: distortion-driven densification ($D^3$) and content-aware Gaussian filters (CAF) to 2D GS with different factorized forms. For GaussinImage, we gradually add our  $D^3$ and CAF into their Cholesky version and RS  variant. When adding $D^3$, we utilize the same sparse initialization $N_0=\frac{M}{2}$ as GaussianImage++. For LIG, we evaluated their two-stage version, which is the best variant, as their paper. We only apply our densification and CAF to the fitting in the second stage.
\subsection{More Analysis Results}
\noindent\textbf{ Runtime analysis  of densification.}
Densification only introduces some computational overhead and tuning complexity during training/encoding rather than the rendering/decoding, ensuring no performance drops in rendering/decoding speed. As shown in \Cref{tab:runtime}, our densification achieves significant quality improvements (1.9 dB in PSNR) with an acceptable increase ( $\sim $16 s ) in training latency.

\begin{table}[htb]
  \centering
  
  \resizebox{0.65\linewidth}{!}{
    \begin{tabular}{lcc}
      \toprule
      5k GS & PSNR & encode(s)  \\
      \midrule
      w / o densification & 29.85  & 102.78 \\ w / growing & 31.71 & 108.65 \\
      w / pruning & 29.89 & 104.52 \\ all & \cellcolor[HTML]{FFCCCC}31.75  & 118.52 \\
      \bottomrule
    \end{tabular}
  }
  \caption{Runtime and complexity analysis about densification.}
  \label{tab:runtime}
\end{table}
\noindent\textbf{ Hyperparameters influence of densification.}
As shown in \Cref{tab:hyperparameters}, smaller pruning step ($t_p \downarrow$) and growing ($t_g \downarrow$) step contribute to image quality, as they frequently pruning invalid GS and densify more GS to a high-frequency area. Besides $t_p$ and $t_g$, our densification involves no other thresholds settings.
\begin{table}[htb]
    \centering
   
    \resizebox{\linewidth}{!}{%
    \begin{tabular}{cccc}
        \toprule
        $t_p$  / $t_g$    & PSNR / MS-SSIM & $t_p$        / $t_g$    & PSNR / MS-SSIM  \\
        \midrule
        100       / 5000     & \cellcolor[HTML]{FFCCCC}31.83 / 0.9599& 100 / 10000    & 31.75 / 0.9594  \\

        500 / 5000     & 31.64 / 0.9586  & 500 / 10000    & 31.56 / 0.9580 \\
        1000        /5000     & 31.69 / 0.9595  &        1000         /10000    & 31.63 / 0.9587 \\
        \bottomrule
    \end{tabular}%
    }
     \caption{Impact of hyperparameters in densification.  }
    \label{tab:hyperparameters}
\end{table}
\begin{table*}[htb]    
    \centering
    \resizebox{\linewidth}{!}{%
        \begin{tabular}{lccclccc}
            \toprule

             $ \bm{\mu}/\mathbf{\Sigma} / \mathbf{c}$ & {BD-PSNR} $\uparrow$ & {BD-Rate (\%)} $\downarrow$ & {BD-MS-SSIM} $\uparrow$ & 
            $ \bm{\mu}/\mathbf{\Sigma} / \mathbf{c}$ & {BD-PSNR} $\uparrow$ & {BD-Rate (\%)} $\downarrow$ & {BD-MS-SSIM} $\uparrow$ \\
            \midrule
               12 / 10 / 6     &     0 &     0     &    0       &  16 / 10  / 6     &    -0.21 &	9.68 & -0.0305   \\
                12 / 10  / 8     &           -0.17 &	3.60 	& -0.0021 & 16 / 10 / 8     &    -0.89 	& 30.85&	-0.0197\\
             12 / 8   / 8     &           -0.72 &	29.42 &	-0.0055 & 16 / 8  / 8     &          -1.03&	44.20 &	-0.0123 
              \\
            12 / 8  / 6     &             -0.49 	& 19.84 &	0.0020  & 16 / 8  / 6     &            -0.82 &	35.11 &-0.0077 
              \\
            \bottomrule
        \end{tabular}
    }
       \caption{Ablation of bit-depth configuration.}
    \label{tab:bitdepth}
\end{table*}  

\begin{table}[htb]
    \centering
    \resizebox{\linewidth}{!}{
        \begin{tabular}{c|ccc}
            \toprule
            & Bpp & PSNR & MS-SSIM \\  
            \midrule
            Mean      & 0.85      & 31.14    & 0.9577   \\  
            Standard  & $3.21 \times 10^{-6}$ & $1.97 \times 10^{-3}$ & $9.03 \times 10^{-7}$ \\  
            \bottomrule
        \end{tabular}
    }
    \caption{Randomness influence of the proposed method for compression results on the Kodak dataset.}
    \label{tab:random}
\end{table}
\noindent\textbf{ Bit-depth configuration analysis of attributes quantization.}
Given a collection of 2D Gaussian primitives, we apply distinct bit-depth LSQ+~\cite{bhalgat2020lsq+} quantizers to different attributes.  
Since geometry attributes ($ \bm{\mu}$ and $\mathbf{\Sigma}$)  determine the dependencies of the rasterization process (eg, the coverage range of the Gaussians), they are more sensitive to deviations than color attributes. Moreover, partial Gaussian primitives possibly become invalid since their de-quantized $\hat{\mathbf{\Sigma}}$  disagrees with the positive definite constraint. So we also periodically pruned them like in the representation task with 100 iterations. Frequent pruning caused by quantization errors will bring performance drops. Hence, we select 10-bit high precision for covariance quantization. For the variance component of $\mathbf{\Sigma}$,  we take the log-uniform quantization due to its uniform distribution. We respectively adopt 12 and 6-bit quantization precision for position and color. ~\Cref{fig:supple_ablation} shows the superiority of our quantization strategies in GaussianImage++ over other different quantization strategies for position and color attributes. 
As shown in \Cref{tab:bitdepth}, our bit-depth configuration (12/10/6) is set to balance bitrate and image quality.
\begin{figure*}[!htb]    
    \centering
    \includegraphics[width=0.75\textwidth]{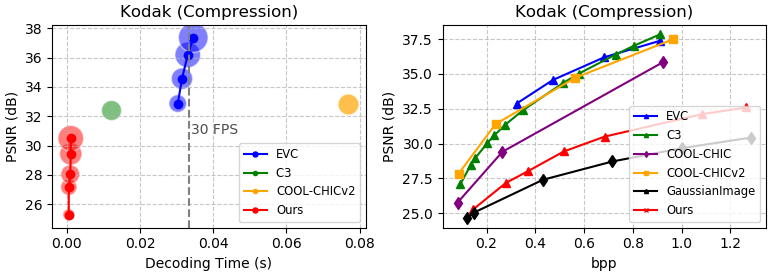}
    \caption{Codec Comparison. The radius (left) of each point
indicates the bpp.}
    \label{fig:comparison_results}
\end{figure*}  

\noindent\textbf{Randomness influences.}  \Cref{tab:random} shows the mean and standard deviation of our method across different random seeds. The tiny standard values indicate that the randomness has a negligible impact on the final results.

\subsection{More Comparison Results
}
\noindent\textbf{Quantitative comparison  with other advanced codecs.
}   GS-based model is a relatively new and promising branch of methods for image compression which currently has a much faster decoding speed but inferior RD compared with SOTA.Consequently,GaussianImage++ mainly outperforms the SOTA VAE-based models, INR-based models and traditional models on decoding speed, model parameters, and low memory usage, as mentioned in the main text.Moreover, GaussianImage++ has demonstrated significant improvement over previous GS methods, including GaussianImage and LIG.  As shown in \Cref{fig:comparison_results}, \textbf{(1) Compared with current SOTA},  our RD performance is not high. It might be too early to require SOTA RD for GS-based methods. \textbf{(2) Compared with Real-time decoding image codecs }, like EVC \cite{wang2023evc}, we also outperform them in decoding speed with inferior RD. Note that our focus is not real-time decoding but to advance this recently proposed 2D GS paradigm for image compression; our method is much faster than real-time ones.

\begin{figure*}[htb]
\centering
\includegraphics[width=\textwidth]{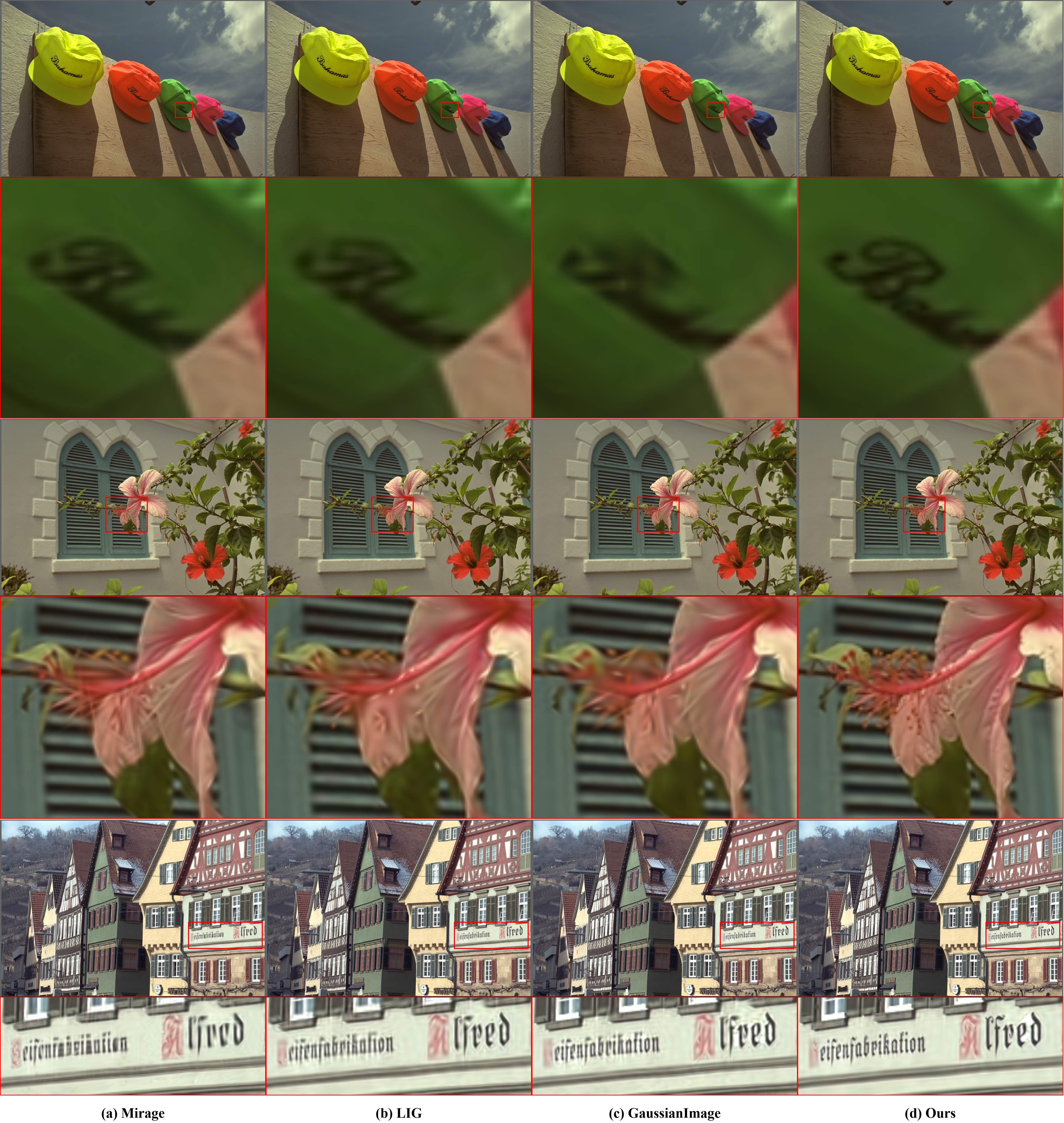}
    \caption{
Qualitative comparison with previous GS-based image representations on the Kodak dataset. (Set total iterations $T=50,000$ and maximum numbers of Gaussians $M=10,000$)
}
\label{fig:supplement_repre}
\end{figure*}

\noindent\textbf{Qualitative comparison. 
}\Cref{fig:supplement_repre} provides a qualitative comparison between our GaussianImage++ and other GS-based methods, including Mirage, LIG and GaussianImage. It reveals that our method restores image details more effectively and delivers superior reconstruction quality with the same number of Gaussians.



\end{document}